# *Kinematic Control of compliant serial manipulators composed of dual-triangles*

*Wanda Zhao*
Laboratoire des Sciences du Numérique de Nantes
Ecole Centrale Nantes
Nantes, France
Wanda.Zhao@ls2n.fr

*Anatol Pashkevich*
Laboratoire des Sciences du Numérique de Nantes
IMT Atlantique Nantes
Nantes, France
Anatol.Pashkevich@imt-atlantique.fr

*Alexandr Klimchik*
Innopolis University
Tatarstan, Russia
A.Klimchik@innopolis.ru

*Damien Chablat*
Laboratoire des Sciences du Numérique de Nantes
Centre National de la Recherche Scientifique (CNRS)
Nantes, France
Damien.Chablat@cnrs.fr

*Abstract*—The paper focuses on the kinematics control of a compliant serial manipulator composed of a new type of dual-triangle elastic segments. Some useful optimization techniques were applied to solve the geometric redundancy problem, ensure the stability of the manipulator configurations with respect to the external forces/torques applied to the end-effector. The efficiency of the developed control algorisms is confirmed by simulation.

*Keywords – compliant manipulator; tensegrity mechanisms; kinematic control; geometric redundancy; elastostatic behavior.*

## I. INTRODUCTION

Compliant serial manipulators attracted much attention in robotics in recent years [1][2][3]. Numbers of novel mechanisms appeared recently, which include not only rigid components but also elastic ones, allowing to achieve excellent flexibility and ability of shape-changing in accordance with the environment. One of the promising trends in this area is using tensegrity mechanisms, whose important advantages are simple design and low weight. However, because of the geometric redundancy and complicated elastostatic properties, the kinematics control of such manipulators is not simple and requires the solution of some theoretical problems considered in this paper.

Robotic manipulators are normally classified into three types conventional discrete, serpentine, and continuum robots [4][5][6][7], the typical earlier hyper-redundant robot designs can be date back to 1970s [8][9]. While designing such a manipulator, researchers are inclined to use a series of similar segments. Relevant studies based on the tensegrity mechanisms focus on the compressive elements and tensile elements (cables or springs). To achieve the desired configurations while working, the manipulators must avoid reaching the unstable equilibriums, but as the number of the mechanism segments increase, the kinematic analysis and control are more and more difficult [10][11][12][13].

A new type of compliant tensegrity mechanism was proposed in our previous papers [14][15]. It is composed of two rigid triangle parts, which are connected by a passive joint in the center and two elastic edges on each side with controllable preload. In this paper, we study a compliant serial manipulator composed of the dual-triangle segments mentioned above, concentrate on the redundancy problem based on the kinematic analysis. Relevant results will be a good base for further investigation.

## II. MECHANICS OF DUAL-TRIANGLE MECHANISM

Let us consider first a single segment of the total serial manipulator to be studied, which consists of two rigid triangles connected by a passive joint whose rotation is constrained by two linear springs as shown in Fig. 1. It is assumed that the mechanism geometry is described by the triangle parameters $(a_1, b_1)$ and $(a_2, b_2)$, and the mechanism shape is defined by the angle that can be adjusted by means of two control inputs influencing on the spring lengths $L_1$ and $L_2$. Let us denote the spring lengths in the non-stress state as $L_1^0$ and $L_2^0$, and the spring stiffness coefficients $k_1$ and $k_2$.

To find the mechanism configuration angle $q$ corresponding to the given control inputs $L_1^0$ and $L_2^0$, let us derive first the static equilibrium equation. From Hook's law, the forces generated by the springs are $F_i = k_i(L_i - L_i^0)$ ($i =1, 2$), where $L_1$, $L_2$ are the spring lengths |AD|, |BC| corresponding to the angle $q$. These values can be computed using the formulas $L_i(\theta_i) = \sqrt{c_1^2 + c_2^2 + 2c_1c_2 \cos(\theta_i)}$ ($i =1, 2$). Here $c_i = \sqrt{a_i^2 + b_i^2}$ ($i =1, 2$), and the angles $\theta_1$, $\theta_2$ are expressed via the mechanism parameters as $\theta_1 = \beta_{12} + q$, $\theta_2 = \beta_{12} - q$, $\beta_{12} = \mathrm{atan}(a_1/b_1) + \mathrm{atan}(a_2/b_2)$. The torques $M_1 = F_1 \cdot h_1$, $M_1 = F_2 \cdot h_2$ in the passive joint O can be computed from the geometry, so we can get

$$\begin{aligned} M_1(q) &= +k_1(1 - L_1^0/L_1(\theta_1))\, c_1 c_2 \sin(\theta_1) \\ M_2(q) &= -k_2(1 - L_2^0/L_2(\theta_2))\, c_1 c_2 \sin(\theta_2) \end{aligned}, \quad (1)$$

where the difference in signs is caused by the different direction of the torques generated by the forces $F_1$, $F_2$ with respect to the passive joint. Further, taking into account the external torque $M_{\mathrm{ext}}$ applied to the moving platform, the static equilibrium equation for the considered mechanism can be written as $M_1(q) + M_2(q) + M_{ext} = 0$.

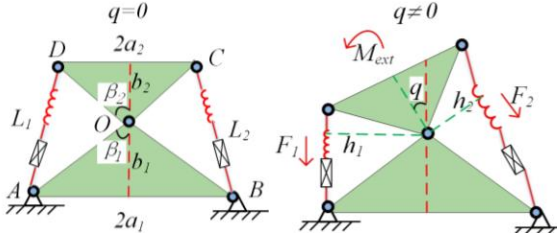

Figure 1. Geometry of a single dual-triangle mechanism.

Let us now evaluate the stability of the mechanism under consideration. In general, this property highly depends on the equilibrium configuration defined by the angle $q$, which satisfies the equilibrium equation $M(q) + M_{ext} = 0$. As follows from the relevant analysis, the function $M(q)$ can be either monotonic or non-monotonic one, so the single-segment mechanism may have multiple stable and unstable equilibriums, which are studied in detail [14][15]. As Fig. 2 shows, the torque-angle curves $M(q)$ that can be either monotonic or two-model one, the considered stability condition can be simplified and reduced to the derivative sign verification at the zero point only, i.e. $M'(q)_{q=0} < 0$, which is easy to verify in practice. It represents the mechanism equivalent rotational stiffness for unloaded configuration with $q=0$.

For the symmetrical case, when $a_1=a_2=a$, $b_1=b_2=b$, $k_1=k_2$, $L_i^0 = L^0$, expressions (1) can be essentially simplified and investigated in detail. It can be proved that relevant torque-angle relations can be either monotonic or non-monotonic, as presented in Fig. 2. It can be also shown that the monotonicity condition can be expressed as follows [14][15]

$$L^0 > 2b \cdot (1 - (a/b)^2), \quad (2)$$

which is easy to verify, and will be used in the following sections to ensure the stability of the kinematic control.

### III. KINEMATICS CONTROL OF MANIPULATOR

As follows from the mechanism structure (see Fig. 1), the desired configuration is defined by a single variable $q$ which is adjusted by two control variables $L_1^0$ and $L_2^0$. The latter creates redundancy and ambiguity in control inputs selection. To eliminate this difficulty, it is reasonable to define $L_1^0$ and $L_2^0$ in a symmetrical way, i.e. as $L_1^0 = L^0 - \Delta$ and $L_2^0 = L^0 + \Delta$. This allows us to write the static equilibrium equation for every single segment as follows

$$M_q = 2k\left[(b^2 - a^2)\sin(q) - L^0 b \sin(q/2) + \Delta \cdot a \cos(q/2)\right], \quad (3)$$

and present the corresponding control law $\Delta(q)$ for the unloaded case ($M_{ext} = 0$) in the following way

$$\Delta(q) = \left[L^0 b \sin(q/2) - (b^2 - a^2)\sin(q)\right] / \left[a \cos(q/2)\right]. \quad (4)$$

It should be noted that the desired configuration defined by the angle $q$ should always satisfy the geometric constraints derived in our previous paper [14][15]. The obtained results are presented in Fig. 3. Also, for the proposed control strategy it is necessary to carefully select initial values of control inputs $L_1^0 = L_2^0$, in order to avoid the negative equivalent rotational stiffness causing instability of the desired configuration of the mechanism.

In a more general case when $M_{ext} \neq 0$, to achieve the desired configuration with the angle $q$ and the external loading $M_{ext}$, the control input $\Delta$ should be computed using a revised expression

$$\Delta(q, M_{ext}) = \frac{M_{ext}/2k + L^0 b \sin(q/2) - (b^2 - a^2)\sin(q)}{a \cdot \cos(q/2)}, \quad (5)$$

which shows that in the loaded case, the symmetrical configuration with $q_0=0$ is achieved by applying a non-zero control input $\Delta$ that compensate the external loading. However, it is necessary to be also careful here about the selection of the parameter $L^0$, which in some cases can cause negative stiffness leading to the buckling phenomenon.

Let us consider now a compliant manipulator composed of three similar segments connected in series as shown in Fig. 4, where the left-hand-side is fixed and the initial configuration is a "straight" one ($q_1=q_2=q_3=0$). This configuration is achieved by applying equal control inputs to all mechanism segments. For this manipulator, to derive the desired control algorism, it is necessary to evaluate the influence of the external force $\mathbf{F}_e=(F_x, F_y)$, which causes the end-effector displacements to a new equilibrium location $(x, y)^T = (6b-\delta_x, \delta_y)^T$ corresponding to the nonzero configuration variables $(q_1, q_2, q_3)$. It is also assumed here the external torque $M_{ext}$ applied to the end-effector is equal to zero. It can be easily proved from the geometry analysis that the configuration angles satisfy the following direct kinematic equations

$$x = b + 2bC_1 + 2bC_{12} + bC_{123}; \quad y = 2bS_1 + 2bS_{12} + bS_{123}, \quad (6)$$

where $C_{123} = \cos(q_1+q_2+q_3)$, $S_{123} = \sin(q_1+q_2+q_3)$, $C_{12} = \cos(q_1+q_2)$, $S_{12} = \sin(q_1+q_2)$, $C_1 = \cos q_1$, $S_1 = \sin q_1$. These two equations include three unknown variables $(q_1, q_2, q_3)$ and allow us to compute two of them assuming that the remaining one is known. For instance, if the angle $q_1$ is assumed to be known, the rest of the angles $q_2$, $q_3$ can be computed from the classical inverse kinematics of the two-link manipulator as follows

$$\begin{aligned} q_3 &= \operatorname{atan}(S_3/C_3) \\ q_2 &= \operatorname{atan}(y - 2bS_1/x - b - 2bC_1) - \operatorname{atan}(bS_3/2b + bC_3) - q_1 \end{aligned}, \quad (7)$$

where $C_3 = \left[(x-b-2bC_1)^2 + (y-2bS_1)^2 - 5b^2\right]/4b^2$, $S_3 = \pm\sqrt{1-C_3^2}$. It is clear that the latter expressions provide two groups of possible solutions corresponding to the positive/negative configuration angles $q_3 \geq 0$ and $q_3 \leq 0$.

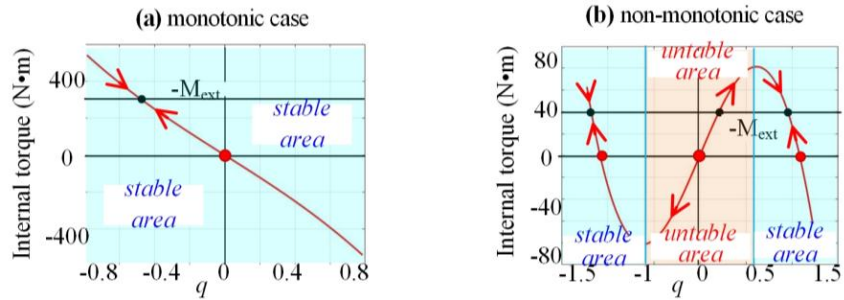

Figure 2. The torque-angle curves and static equilibriums for $L_1^0 = L_2^0$ ($q_0 = 0$).

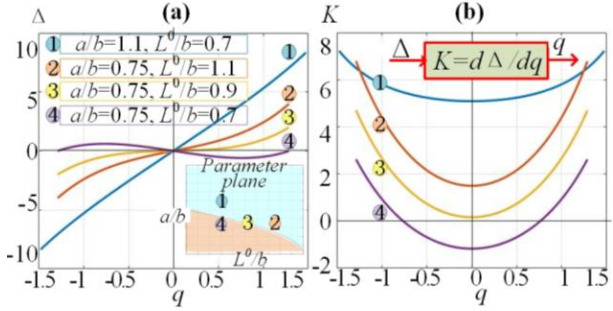

Figure 3.  Relations between the control input $\Delta$, sensitivity coefficient $K$, and the desired configuration angle $q$

To achieve the desired end-point position $(x, y)$, it is clear that there is an obvious redundancy here related to the selection of three configuration angles $(q_1, q_2, q_3)$ allowing to reach the target point described by two Cartesian Coordinates $(x, y)$, but this problem is outside of the stiffness analysis and should be solved using other techniques (obstacle avoidance, minimization of joint motions, etc.) The simplest way to overcome the redundancy problem is to minimize *the joint motions* by moving from the initial configuration $(q_1^o, q_2^o, q_3^o)$ to a final one $(q_1, q_2, q_3)$ corresponding to the desired end-point $(x, y)$. This objective can be expressed formally in several ways, for example as

a)  *Minimization of the total sum of the joint angle increments*: $\sum_{i=1}^{3}|q_i - q_i^o| \to \min$ ;

b)  *Minimization of the largest joint angle increment*: $\max_i |q_i - q_i^o| \to \min$

It is clear that such an optimization problem should be solved with respect to the two scalar constraints arising from (6). The latter gives us a simple numerical technique where the joint angle $q_1$ is an independent variable and angles $q_2$, $q_3$ are computed via the inverse kinematics (7) taking into account the duality expressed by the '$\pm$' sign. An example of this approach is presented in Fig. 5 where the objectives (*a*) and (*b*) give slightly different solutions both of which are acceptable in practice. There is also an alternative approach,

c)  *Minimize the total sum of joint angle increment squares* : $\sum_{i=1}^{3}(q_i - q_i^o)^2 \to \min$ ,

For this objective, if the initial and target points are close enough, we can apply linearization and express the direct kinematic constraints in the form of two linear equations

$$\Delta \mathbf{p} = \left[\mathbf{J}_{ij}\right]_{2\times 3} \cdot \Delta \mathbf{q} , \quad (8)$$

where

$$\left[\mathbf{J}_{ij}\right] = \begin{bmatrix} -2bS_1 - 2bS_{12} - bS_{123} & -2bS_{12} - bS_{123} & -bS_{123} \\ 2bC_1 + 2bC_{12} + bC_{123} & 2bC_{12} + bC_{123} & bC_{123} \end{bmatrix} \quad (9)$$

$\Delta \mathbf{p} = (x-x_0, y-y_0)^T$ and $\Delta \mathbf{q} = (q_1 - q_1^o, q_2 - q_2^o, q_3 - q_3^o)^T$, Such approach leads the following constraint optimization problem: minimize the function of the joint angle increments $\Delta q_1, \Delta q_2, \Delta q_3$

$$f(\Delta q_1, \Delta q_2, \Delta q_3) @ \left(\Delta q_1^2 + \Delta q_2^2 + \Delta q_3^2\right)/2 \to \min , \quad (10)$$

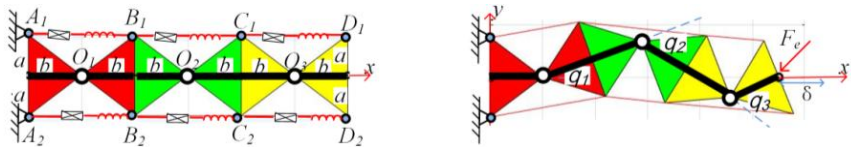

Figure 4.  The torque-angle curves and static equilibriums for $L_1^0 = L_2^0$ ($q_0 = 0$).

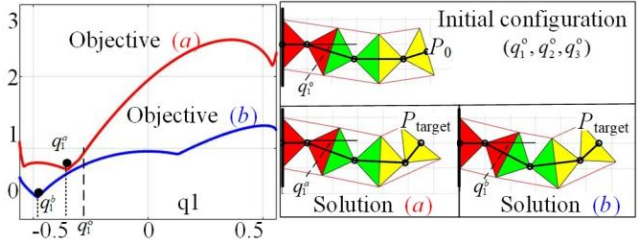

Figure 5. Kinematic control of a redundant manipulator via minimization of objectives (*a*) and (*b*).

subject to the equality constraints

$$g_1(\Delta q_1, \Delta q_2, \Delta q_3) @ \Delta x - (J_{11}\Delta q_1 + J_{12}\Delta q_2 + J_{13}\Delta q_3) = 0$$
$$g_2(\Delta q_1, \Delta q_2, \Delta q_3) @ \Delta y - (J_{21}\Delta q_1 + J_{22}\Delta q_2 + J_{23}\Delta q_3) = 0 \quad . \quad (11)$$

These problems can be solved using the Lagrange technique by minimizing the function of five variables

$$L(\Delta q_1, \Delta q_2, \Delta q_3, \lambda_1, \lambda_2) @ f(.) + \lambda_1 g_1(.) + \lambda_2 g_2(.) \to \min , \quad (12)$$

where $\lambda_1$ and $\lambda_2$ are Lagrange multipliers. Further, after setting to zero the gradient $\nabla L = 0$, which is composed of the partial derivatives, one can obtain the following scalar equations with respect to the variables $q_i$ and $\lambda_j$, ($i = 1, 2, 3$).

$$\Delta q_i - \sum_{j=1}^{2} \lambda_j J_{ji} = 0 \; ; \; \sum_{i=1}^{3} J_{1i} q_i = \Delta x \; ; \; \sum_{i=1}^{3} J_{2i} q_i = \Delta y \; , \quad (13)$$

that can be presented in the matrix form as follows

$$\begin{bmatrix} \mathbf{I}_{3\times 3} & -\mathbf{J}_{3\times 2}^T \\ \mathbf{J}_{2\times 3} & \mathbf{0}_{2\times 2} \end{bmatrix} \cdot \begin{bmatrix} \Delta \mathbf{q} \\ \lambda \end{bmatrix} = \begin{bmatrix} \mathbf{0}_{3\times 1} \\ \Delta \mathbf{p} \end{bmatrix}, \quad (14)$$

where $\lambda = (\lambda_1, \lambda_2)^T$. Using the block matrix inverse, the desired solution can be expressed as

$$\begin{bmatrix} \Delta \mathbf{q} \\ \lambda \end{bmatrix} = \begin{bmatrix} \mathbf{I} - \mathbf{J}^T(\mathbf{J}\mathbf{J}^T)^{-1}\mathbf{J} & \mathbf{J}^T(\mathbf{J}\mathbf{J}^T)^{-1} \\ -(\mathbf{J}\mathbf{J}^T)^{-1}\mathbf{J} & (\mathbf{J}\mathbf{J}^T)^{-1} \end{bmatrix} \cdot \begin{bmatrix} \mathbf{0}_{3\times 1} \\ \Delta \mathbf{p} \end{bmatrix}, \quad (15)$$

which yields the following vector of the joint angle increments

$$\Delta \mathbf{q} = \mathbf{J}^T (\mathbf{J}\mathbf{J}^T)^{-1} \Delta \mathbf{p} \; . \quad (16)$$

It should be noticed that the latter expression is similar to the matrix pseudo-inverse of Moore-Penrose. Besides, to achieve the equilibriums corresponding to the desired configurations, the three-segment mechanism must be controlled by three pairs of the control inputs ($L_{11}^0, L_{12}^0$), ($L_{21}^0, L_{22}^0$) and ($L_{31}^0, L_{32}^0$). To simplify the mechanism control, let us apply the symmetrical approach in previous, which allows using only three control variables ($\Delta_1, \Delta_2, \Delta_3$) producing six physical control inputs $L_{i1}^0 = L^0 - \Delta_i$, $L_{i2}^0 = L^0 + \Delta_i$ with $i = 1, 2, 3$, where the values of $\Delta_i$ are computed using formulas from the one-segment mechanism control law (4). The above-described approach gives us the following algorithm for the control of the three-segment mechanism:

1) Using the direct kinematics equations (6) and additional objectives allowing to resolve the kinematic redundancy, compute the configuration angles $q_1(x, y)$, $q_2(x, y)$ and $q_3(x, y)$ corresponding to the desired end-point position $(x, y)$ and ensuring the manipulator "minimum motions" of the joints.

2) Using expression (4), compute the control inputs $\Delta_1(q_1)$, $\Delta_2(q_2)$ and $\Delta_3(q_3)$ for the three segments corresponding to the configuration angles ($q_1, q_2, q_3$).

An example of computing based on the above algorithm is presented in Fig. 6, where the mechanism parameters $a/b=1.1$, $L^0/b =0.7$ were chosen to ensure the mechanism stability in the unloaded mode $M_{ext}=0$, and the initial configuration is $q_0 = (-0.1, 0.1, 0.1)$.

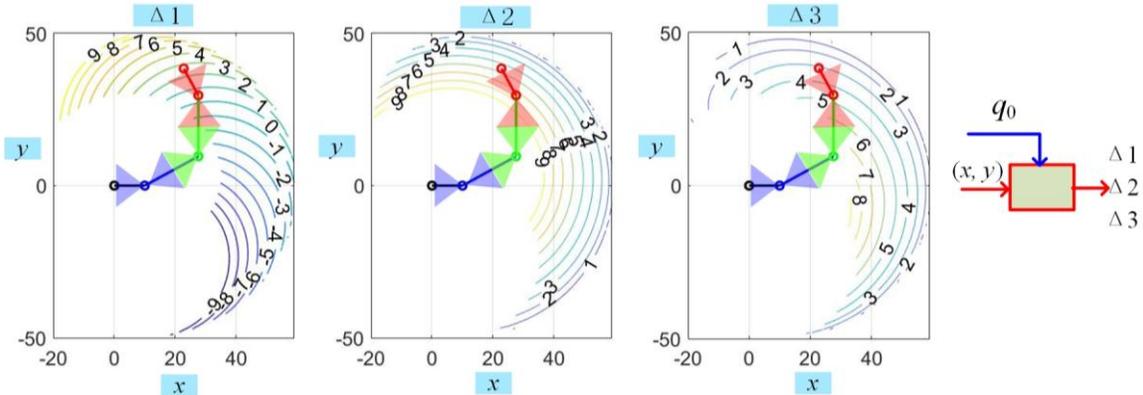

Figure 6. Relations between the control inputs ($\Delta_1, \Delta_2, \Delta_3$) and the desired end-point position $(x, y)$ with an initial configuration $q_0 = (-0.1, 0.1, 0.1)$ and parameters $a/b=1.0$, $L^0/b=1.0$ (unloaded case $F_x = 0$, $F_y = 0$).

In more general case when the external forces ($F_x, F_y$) are not equal to zero, it is also suggested to solve the redundant inverse kinematic problem using the above-presented objectives (*a*), (*b*), (*c*) (i.e. to use the configuration angles $q_1$, $q_2$, $q_3$ from the unloaded case), but to compute the modified control inputs allowing to compensate the external load. The corresponding algorithm implementing this technique is presented below.

1) Using the direct kinematics equations (6) and additional objectives allowing to resolve the kinematic redundancy, compute the configuration angles $q_1(x,y)$, $q_2(x,y)$ and $q_3(x,y)$ corresponding to the desired end-point position $(x, y)$ and ensuring the manipulator "minimum motions" of the joints.
2) Compute the joint torques $(M_{q1}, M_{q2}, M_{q3})$ corresponded to the external force ($F_x, F_y$) applied at the manipulator end-point.
3) Using expression (5), compute the control inputs $\Delta_1(q_1)$, $\Delta_2(q_2)$ and $\Delta_3(q_3)$ for the three segments corresponding to the configuration angles ($q_1$, $q_2$, $q_3$) and the joint torques $(M_{q1}, M_{q2}, M_{q3})$.

It can be demonstrated that such algorism can ensure the stable configurations of the manipulator.

## IV. Conclusion

The paper focuses on kinematic control of the compliant serial manipulator composed of a new type of dual-triangle tensegrity mechanisms, which are composed of rigid triangles connected by passive joints. The manipulator shape is controlled by adjusting the initial lengths of the elastic components located on two edges of each compliant segment. The main difficulties in control of such mechanisms are related to geometric redundancy and complicated behavior under the loading, which may be unstable if the control inputs are not selected properly.

The developed control algorism allows users to compute the control variables, which ensure the end-effector displacement to the desired location using very efficient motion that corresponds to minimum increments of all joint coordinates simultaneously. Besides, during such motion, these control inputs ensuring elastostatic stability of the manipulator shape with respect to the external forces/torques applied on the end-effector. The proposed algorism were carefully investigated via simulation, which confirmed via the results presented in this paper. Further research based on this study will focus on the more complicated manipulator motions in a constrained environment.

## Acknowledgment

This work was supported by the China Scholarship Council ( No. 201801810036 ).

## References

[1] M. I. Frecker, G. K. Ananthasuresh, S. Nishiwaki, N. Kikuchi, and S. Kota, "Topological Synthesis of Compliant Mechanisms Using Multi-Criteria Optimization," Journal of Mechanical Design, vol. 119, no. 2, pp. 238–245, Jun. 1997, doi: 10.1115/1.2826242.

[2] A. Albu-Schaffer et al., "Soft robotics," IEEE Robotics Automation Magazine, vol. 15, no. 3, pp. 20–30, Sep. 2008, doi: 10.1109/MRA.2008.927979.

[3] M. Y. Wang and S. Chen, "Compliant Mechanism Optimization: Analysis and Design with Intrinsic Characteristic Stiffness," Mechanics Based Design of Structures and Machines, vol. 37, no. 2, pp. 183–200, May 2009, doi: 10.1080/15397730902761932.

[4] G. Robinson and J. B. C. Davies, "Continuum robots - a state of the art," in Proceedings 1999 IEEE International Conference on Robotics and Automation (Cat. No.99CH36288C), May 1999, vol. 4, pp. 2849–2854 vol.4, doi: 10.1109/ROBOT.1999.774029.

[5] G. S. Chirikjian and J. W. Burdick, "Kinematically optimal hyper-redundant manipulator configurations," IEEE Transactions on Robotics and Automation, vol. 11, no. 6, pp. 794–806, Dec. 1995, doi: 10.1109/70.478427.

[6] J. Yang, E. P. Pitarch, J. Potratz, S. Beck, and K. Abdel-Malek, "Synthesis and analysis of a flexible elephant trunk robot," Advanced Robotics, vol. 20, no. 6, pp. 631–659, Jan. 2006, doi: 10.1163/156855306777361631.

[7] G. S. Chirikjian and J. W. Burdick, "A modal approach to hyper-redundant manipulator kinematics," IEEE Transactions on Robotics and Automation, vol. 10, no. 3, pp. 343–354, Jun. 1994, doi: 10.1109/70.294209.

[8] Anderson, Victor C., and Ronald C. Horn. "Tensor arm manipulator." U.S. Patent No. 3,497,083. 24 Feb. 1970.

[9] I. A. Gravagne and I. D. Walker, "Kinematic transformations for remotely-actuated planar continuum robots," in Proceedings 2000 ICRA. Millennium Conference. IEEE International Conference on Robotics and Automation. Symposia Proceedings (Cat. No.00CH37065), Apr. 2000, vol. 1, pp. 19–26 vol.1, doi: 10.1109/ROBOT.2000.844034.

[10] R. E. Skelton and M. C. de Oliveira, Tensegrity systems. Berlin: Springer, 2009.

[11] K. W. Moored, T. H. Kemp, N. E. Houle, and H. Bart-Smith, "Analytical predictions, optimization, and design of a tensegrity-based artificial pectoral fin," International Journal of Solids and Structures, vol. 48, no. 22–23, pp. 3142–3159, Nov. 2011, doi: 10.1016/j.ijsolstr.2011.07.008.

[12] M. Arsenault and C. M. Gosselin, "Kinematic, static and dynamic analysis of a planar 2-DOF tensegrity mechanism," Mechanism and Machine Theory, vol. 41, no. 9, pp. 1072–1089, Sep. 2006, doi: 10.1016/j.mechmachtheory.2005.10.014.

[13] M. Furet, M. Lettl, and P. Wenger, "Kinematic Analysis of Planar Tensegrity 2-X Manipulators," in Advances in Robot Kinematics 2018, vol. 8, J. Lenarcic and V. Parenti-Castelli, Eds. Cham: Springer International Publishing, 2019, pp. 153–160.

[14] Zhao, W., Pashkevich, A., Klimchik, A. and Chablat, D., "Stiffness Analysis of a New Tensegrity Mechanism based on Planar Dual-triangles". In Proceedings of the 17th International Conference on Informatics in Control, Automation and Robotics - Vol 1: ICINCO, July. 2020, ISBN 978-989-758-442-8, pages 402-411. doi: 10.5220/0009803104020411

[15] Zhao, W., Pashkevich, A., Klimchik, A. and Chablat, D., "The Stability and Stiffness Analysis of a Dual-Triangle Planar Rotation Mechanism". In Proceedings of ASME 2020 International Design Engineering Technical Conferences And Computers and Information in Engineering Conference IDETC/CIE, Aug. 2020, (in press).